\documentclass[conference]{IEEEtran}\IEEEoverridecommandlockouts
\usepackage{cite}
\usepackage{amsmath,amssymb,amsfonts}
\usepackage{algorithmic}
\usepackage{graphicx}
\usepackage{textcomp}
\usepackage{xcolor}
\usepackage{xcolor}
\usepackage{graphicx}
\usepackage{verbatim}
\usepackage{array}
\usepackage{tabularx}
\usepackage{makecell}
\usepackage{caption}
\usepackage{float}
\usepackage{subfig}
\usepackage{amsmath}
\usepackage{booktabs}
\usepackage{threeparttable} 
\usepackage{color}
\usepackage{mathtools}

\usepackage{mwe}
\usepackage{tabularx}
\usepackage{tikz}
\usepackage{pgfplots}
\usetikzlibrary{patterns}
\def\BibTeX{{\rm B\kern-.05em{\sc i\kern-.025em b}\kern-.08em
    T\kern-.1667em\lower.7ex\hbox{E}\kern-.125emX}}
\begin{document}

\title{Graph-Augmented LLMs for Personalized Health Insights: A Case Study in Sleep Analysis}

\author{Ajan Subramanian$^{1,*}$,  Zhongqi Yang$^{1}$, Iman Azimi$^{1}$, and Amir M. Rahmani$^{1,2}$ 
\thanks{$^{1}$Dept. of Computer Science, University of California, Irvine. 
        $^{2}$School of Nursing, University of California, Irvine. 
        {\tt\small (*correspondence email: ajans1@uci.edu)}}%
}

\maketitle

\begin{abstract}
Health monitoring systems have revolutionized modern healthcare by enabling the continuous capture of physiological and behavioral data, essential for preventive measures and early health intervention. While integrating this data with Large Language Models (LLMs) has shown promise in delivering interactive health advice, traditional methods like Retrieval-Augmented Generation (RAG) and fine-tuning often fail to fully utilize the complex, multi-dimensional, and temporally relevant data from wearable devices. These conventional approaches typically provide limited actionable and personalized health insights due to their inadequate capacity to dynamically integrate and interpret diverse health data streams. In response, this paper introduces a graph-augmented LLM framework designed to significantly enhance the personalization and clarity of health insights. Utilizing a hierarchical graph structure, the framework captures inter and intra-patient relationships, enriching LLM prompts with dynamic feature importance scores derived from a Random Forest Model. The effectiveness of this approach is demonstrated through a sleep analysis case study involving 20 college students during the COVID-19 lockdown, highlighting the potential of our model to generate actionable and personalized health insights efficiently. We leverage another LLM to evaluate the insights for relevance, comprehensiveness, actionability, and personalization, addressing the critical need for models that process and interpret complex health data effectively. Our findings show that augmenting prompts with our framework yields significant improvements in all 4 criteria. Through our framework, we can elicit well-crafted, more thoughtful responses tailored to a specific patient. 

\end{abstract}

\section{Introduction}

\begin{figure*}[h!]
    \centering
    \includegraphics[width=\textwidth]{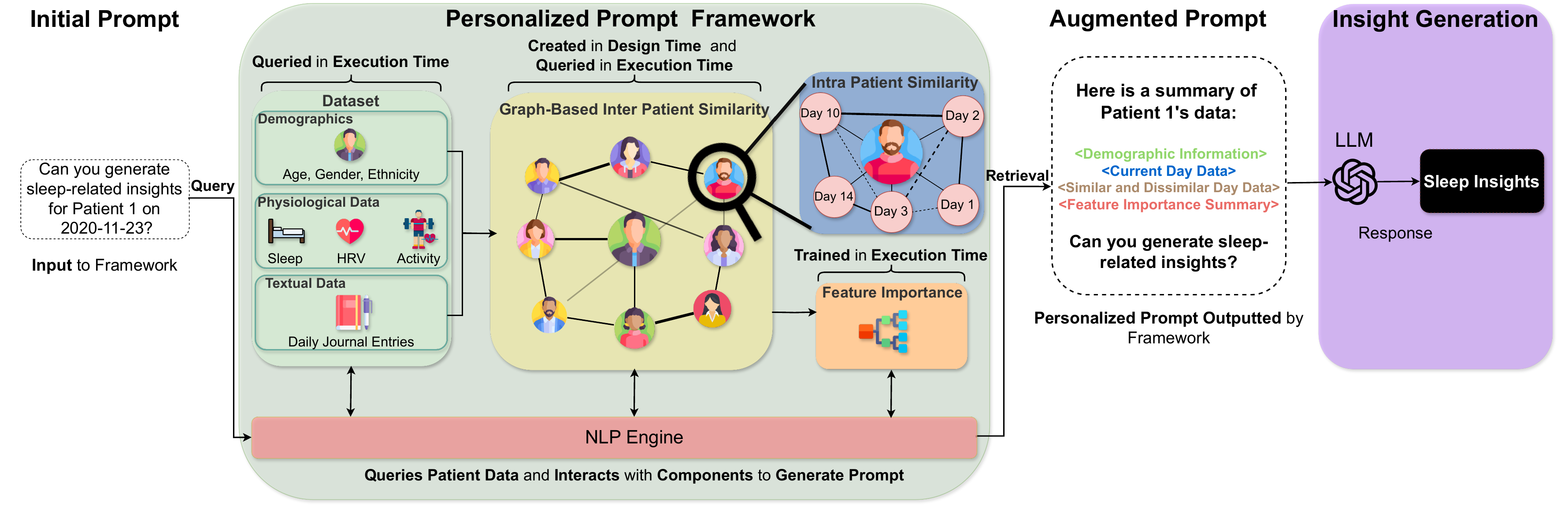}
    \caption{Graph-Based Personalized Prompt Framework with Sleep Analysis as a Case Study}
    \label{fig:per_rag}
\end{figure*}

Wearable health monitoring systems are emerging in modern healthcare, enabling continuous tracking and analysis of physiological and behavioral data in everyday settings~\cite{sano2018identifying}. These systems provide various health parameters, such as heart rate, activity levels, and sleep patterns. Recent advancements in mobile health and wearable sensing technologies have further enhanced the development of machine learning models to extract relationships between various objective well-being measures~\cite{yang2024integrating,yang2023loneliness, jafarlou2023objective}. 
These technologies facilitate the objective measurement of sleep patterns and associated physiological features such as heart rate variability, activity levels, behavior, environmental influences, etc. and play a crucial role in early detection and intervention of health conditions. 


Recently, health monitoring frameworks powered by Large Language Models (LLMs) have emerged as a significant advancement, showing promise in providing personalized and explainable question-answering services~\cite{ali2023using,yang2023exploring}.
The potential of those frameworks in healthcare lies in their extensive training on vast corpora and their ability to be augmented with various types of input through fine-tuning, Retrieval-Augmented Generation (RAG) frameworks~\cite{lewis2020retrieval}, or agent-based methods. 
Given their ability to understand and generate responses in natural language, LLMs enhance patient interactions by providing clear and understandable responses through these frameworks.

Several pioneering studies have demonstrated the use of LLMs for health monitoring, particularly when augmented by wearable sensors~\cite{liu2023using,kim2024health,yang2024chatdiet,abbasian2024knowledge,abbasian2023conversational,yang2023exploring,cosentino2024personal}. 
Generally, these approaches can be categorized into two main groups.
The first group directly prompts raw signal values from wearable devices to LLMs to execute tasks. 
For example, Health-LLM transcribes time series signals from wearable devices and inputs them into LLMs to enhance forecasting capabilities~\cite{kim2024health}.
The second category involves an intermediate step to process data and convert it into natural language text. This step may use statistical models, machine learning models, or other techniques designed to process data more efficiently than LLMs~\cite{yang2024chatdiet,abbasian2024knowledge,merrill2024transforming}. 
The processed outputs are then transcribed into text and fed into LLMs using RAG or agent-based approaches.
For example, Chatdiet demonstrates the augmentation of LLMs with causal inference on wearable data to provide personalized dietary recommendations~\cite{yang2024chatdiet}. Similarly, \cite{abbasian2024knowledge} employs an agentic framework, OpenCHA, which integrates external augmentations organized by an orchestrator to improve diabetes management~\cite{abbasian2023conversational}.
These advancements indicate the potential of merging wearable technology with language models to develop cutting-edge health monitoring solutions.

However, the existing approaches still fall short in extracting complex insights from the data collected by wearable and mobile devices.
These methods struggle to provide high-quality personalized responses due to their inability to fully leverage the multi-dimensional and temporally relevant data~\cite{spathis2023first}. 
Directly inputting raw signal values from wearable sensors hinders the LLMs' ability to understand these signals, as LLMs are not designed or trained to handle numerical data effectively. 
The second category of methods heavily depends on intermediate data processing models, but these models are not optimized for effective integration with LLMs. As a result, they struggle to capture comprehensive health insights from wearables, particularly temporal patterns and inter-patient similarities.
There is a critical need for enhanced models that can effectively process and interpret complex personal health data to offer clear, explainable, and actionable insights.


In this paper, we introduce a graph-augmented LLM framework to generate personalized health insights. The framework utilizes a hierarchical graph structure, where each patient is a node and each day's data are sub-nodes, connected by cosine similarity to reflect daily and overall pattern similarities. This setup allows the graph to enhance LLM prompts with feature importance scores from a Random Forest Model trained on historical and similar patients' data. We assess the framework's effectiveness in a case study focused on sleep analysis, using data from 20 college students in Southern California during the COVID-19 lockdown in March 2020. We evaluate the quality of our framework's insights by employing another LLM to rate these responses on four main criteria - relevance, comprehensiveness, actionability, and personalization. 

\section{Method}

We propose a graph-augmented LLM framework to deliver personalized health insights by enhancing user prompts with patient data. It allows users to query a patient’s wearable data, such as sleep patterns, activity levels, and heart rate, for particular dates. The process starts with an NLP engine that extracts essential details like patient ID, metrics of interest, and date from the prompt. The engine retrieves related demographic and date-specific data from a dataset. For personalization, we develop an inter and intra-patient similarity graph to depict hierarchical data relationships. The NLP engine interacts with the graph to identify the most relevant dates relative to the query and supports a feature importance model that determines key factors affecting health metrics based on similar patient data. The resulting augmented prompt, enriched with these insights, is fed to an LLM to generate personalized health insights. An overview of the framework architecture is indicated in Figure \ref{fig:per_rag}. In the following, we outline each component of the proposed framework.

\subsection{NLP Engine}

The NLP engine coordinates user queries, extracting key identifiers like patient ID and date from the prompt. It accesses the dataset to retrieve relevant demographic and date-specific data, aligning the augmented prompt with the user's query. Utilizing a pre-established inter and intra-patient similarity graph, the engine identifies pertinent temporal and patient comparison patterns. It also integrates feature importance scores from a model trained on data from similar patients. This integration enriches the user's initial prompt with comprehensive contextual details, formatted for optimal understanding by an LLM. 

\subsection{Dataset}
The dataset comprises de-identified patient records, including demographic, wearable, and textual data, indexed by patient ID and date. The NLP engine accesses this dataset to retrieve relevant patient-specific information using a Retrieval-Augmented Generation (RAG) approach \cite{lewis2020retrieval}.

\subsection{Inter and Intra-patient Similarity Graph}
The graph integrates diverse physiological, behavioral, and textual data, utilizing NLP techniques like sentiment analysis to extract sentiments and themes relevant to desired health outcomes. After careful selection and standardization, these multimodal data points form a comprehensive, multidimensional patient profile. We construct an inter and intra-patient similarity graph using these profiles, employing cosine similarity to quantify and link relationships based on similarities and differences among days and patients. This structure is essential for dynamically refining LLM inputs, thereby enhancing its ability to generate precise health insights by recognizing patterns across patient data.

\subsection{Feature Importance}
We develop an ML model (i.e., Random Forest) to identify the most influential factors affecting the health metric of interest, using historical data and information from patients most similar derived from the graph. This model, crucial for pinpointing key variables, is trained with data extracted from the graph during each user prompt. Details from the prompt like the patient's ID and specified date guide the NLP engine in extracting relevant information from the graph. The derived feature importance scores are then used to dynamically enhance the prompt template, improving the accuracy and relevance of the insights generated.

\subsection{Insight Generation}
In the final stage of our pipeline, we employ an LLM to generate personalized health insights. The prompts are enriched with relevant contextual information from the patient’s profile and their network of similar patients. The LLM leverages its advanced natural language capabilities and extensive pre-trained knowledge to provide nuanced insights into individual health patterns and well-being. This method combines the insights generated from traditional machine learning with the reasoning abilities and vast pre-trained knowledge of advanced NLP to produce comprehensive, personalized health recommendations tailored to each patient.

\section{Case Study: Sleep Analysis}
We evaluate our framework through a sleep analysis case study involving 20 college students collected during the 2020 COVID-19 pandemic. Below, we outline the dataset in detail and the integrated setup of our framework's components.

\subsection{Dataset}
We utilize data from a comprehensive remote health monitoring study, capturing physiological and psychophysiological parameters using the Oura rings \cite{OuraRing} and Samsung Galaxy Active 2 smartwatches. Each patient had data for an average duration of 7.8 months (SD=3.8, MIN=1.0, MAX=13.4). The ring provided detailed metrics on sleep stages, duration, and quality, as well as daily activity levels and cardiovascular health indicators. Additionally, we incorporated demographic information (age, gender, ethnicity) and daily diary entries collected through ecological momentary assessments (EMA) \cite{jafarlou2023objective, labbaf2023zotcare}. Details of the data can be found in this article \cite{labbaf2024physiological}.

\subsection{Setup}

\textbf{NLP Engine:} Our NLP engine was implemented using Python libraries including Pandas, Re (regex), and FuzzyWuzzy \cite{Cohen2011}, which processes user prompts into commands that query the dataset and augment the prompts for deeper analysis.

\textbf{Graph Construction and Feature Importance:}
To construct our graph, we use measurements relevant to sleep such as sleep duration, wakefulness, REM stages, sleep quality scores, heart rate variability (HRV), and activity levels. Complementing this, we analyze textual data from daily journal entries using the \textbf{`transformers’} library, specifically employing Facebook’s BART MNLI model \cite{shu2022zero} within a zero-shot classification framework to extract sentiments and six broad themes spanning academics, personal well-being, and social interactions. We extract feature importance scores for sleep score based on information extracted from the graph during query execution. We use a Random Forest model with 100 trees. 

\textbf{Insight Generation:}
We use GPT-4 from the OpenAI API \cite{achiam2023gpt} to generate insights. GPT-4’s vast pre-trained knowledge and in-depth reasoning capabilities allow it to produce well-crafted responses that are easily accessible to non-experts.

\section{Results}

\begin{table*}[htbp]
\centering
\begin{threeparttable}
  \caption{Average Scores (out of 10) of LLM Evaluations Across Different Stages}
  \label{tab:average_scores}
  \begin{tabular}{|l|c|c|c|c|c|}
    \hline
    \textbf{Stage} & \textbf{Relevance} & \textbf{Comprehensiveness} & \textbf{Actionability} & \textbf{Personalization} & \textbf{Uses Graph} \\ \hline
    Demographic Information & $7.61 \pm 1.19$ & $6.80 \pm 1.15$ & $6.10 \pm 1.76$ & $3.61 \pm 1.34$ & No \\
    Current Day Information & $8.65 \pm 0.58$ & $7.48 \pm 0.68$ & $6.45 \pm 0.88$ & $6.08 \pm 1.12$ & No \\
    Similar/Dissimilar Days Info & \textbf{8.73 $\pm$ 0.50} & \textbf{7.51 $\pm$ 0.66} & \textbf{6.51 $\pm$ 0.84} & $6.27 \pm 1.18$ & \textbf{Yes} \\
    Feature Importance & $8.66 \pm 0.64$ & $7.43 \pm 0.65$ & $6.47 \pm 0.69$ & \textbf{6.38 $\pm$ 1.17} & \textbf{Yes} \\
    \hline
  \end{tabular}
  \begin{tablenotes}
    \small
    \item \textbf{Note}: Each stage is incremental in that it adds new information to the previous stage.
  \end{tablenotes}
\end{threeparttable}
\end{table*}

In this section, we assess the effectiveness of our proposed framework in delivering personalized sleep insights given a patient's profile. Our objective is to determine the framework’s ability to create a personalized prompt that would elicit a nuanced, well-crafted response from the LLM. We compare the performance of our framework by incrementally refining the input prompts supplied to the LLM. We started with basic demographic information, then sequentially added current-day data, followed by similar and dissimilar day information, and finally the feature importance. Each variation was designed to explore how additional contextual data could enhance the LLM’s output quality. 

To evaluate the output quality of the LLM, we utilized another GPT-4 model, to perform initial evaluations of the responses generated. We provided the evaluator LLM with definitions of four criteria—relevance, comprehensiveness, actionability, and personalization—and asked it to rate each criterion on a scale from 0 to 10. This process, termed ``LLM-guided evaluation'' simulates a peer-review mechanism within a controlled experimental setup, leveraging the advanced natural language understanding capabilities of LLMs \cite{arthur2023llm}. Using an LLM as a ``reviewer'' was instrumental in providing a scaleable and consistent preliminary assessment tool, enabling us to systematically analyze the quality of outputs. 

We conducted this incremental analysis using 50 diverse prompts across all patients in our dataset. To minimize any implicit bias from the sequence of data presentation, we first stored the outputs of each stage and then shuffled them before the assessment. The evaluation of these outputs was based on the predefined criteria. We calculated the mean scores and their standard deviations for each stage. The results, along with LLM-generated ratings for each stage, are displayed in Table \ref{tab:average_scores}. Given that both the generation and evaluation of insights are conducted using the same GPT-4 model, the relative differences in the scores across stages are more indicative of the framework's incremental improvements than their absolute values.

The results shown in Table \ref{tab:average_scores} indicate a notable improvement in the quality of LLM outputs as reflected by the aggregated scores for each of the 4 criteria. 

\begin{itemize}
    \item \textbf{Demographic Information Only}: This initial stage provided a baseline for personalization with scores indicating the least personalized insights. 
    \item  \textbf{Current Day Information}: The NLP engine queries the dataset to obtain information from both this stage and the previous one. The addition of current-day data improves all metrics. This immediate date-specific context allows the LLM to produce insights that are significantly more tailored and actionable because they directly address the patient's most recent conditions. 
    \item  \textbf{Similar/Dissimilar Days information}: The NLP engine queries the similarity graph to obtain this information. Introducing comparisons between similar and dissimilar days further enriched the model's outputs. This stage had the best scores across most categories showing that the comparative analysis helps the LLM identify patterns or anomalies over time, providing deeper insights into the patient's health fluctuations. 
    \item  \textbf{Feature Importance}: The NLP engine extracts this information from the feature importance component of our framework. The most detailed stage incorporated the integration of feature importance metrics. This increased personalization. This stage obtains feature importance from the patient's history and their nearest neighbors. The personalization scores are the highest in this stage highlighting the LLM's ability to focus on the most impactful health factors that are specific to the patient. 
    
\end{itemize}

Our incremental enhancement approach revealed that the depth and utility of insights from the LLM improved as more detailed and contextual information was added to the input prompt. Particularly, the introduction of similar and dissimilar day data allowed the LLM to generate more personalized and actionable health insights, reflecting a deeper understanding of patient profiles through a graph-based similarity approach.

\section{Conclusion}

This paper proposed a graph-augmented LLM framework that dynamically integrates complex, multi-dimensional data to enhance the personalization and accuracy of health insights. Utilizing a hierarchical graph, the framework effectively captures inter and intra-patient relationships, enriching LLM prompts for richer, context-aware interactions. The effectiveness of the framework was validated through a sleep analysis case study involving college students during the COVID-19 lockdown. The framework's performance was assessed using a novel LLM-guided evaluation, which evaluated insights on relevance, comprehensiveness, actionability, and personalization, showing significant improvements, especially with the introduction of graph-based similarities. For future work, integrating Graph Neural Networks (GNNs) could further enhance the framework's ability to process and interpret complex data structures, improving the generation of real-time, personalized health insights.

\bibliographystyle{IEEEtran}
\bibliography{IEEEabrv,ref}

\end{document}